\documentclass[final, authoryear,5p]{elsarticle}

\usepackage{lineno,hyperref}
\modulolinenumbers[5]

\journal{*}

\usepackage{amsmath, amssymb}
\usepackage{amsfonts}
\usepackage{algorithm, algorithmic}
\usepackage{stfloats}
\usepackage{multirow}
\usepackage{multicol}
\usepackage{makecell}
\usepackage{pgfplots}
\usepackage{tikz}
\usepackage{tablefootnote}
\usepackage{array}
\usepgfplotslibrary{statistics}
\usetikzlibrary{positioning}
\definecolor{ao}{rgb}{0.0, 0.7, 0.0}
\usepackage{scalefnt}

\newcolumntype{M}{@{\extracolsep{1cm}}c@{\extracolsep{5pt}}}%

\newcolumntype{m}{@{\extracolsep{0.6cm}}c@{\extracolsep{5pt}}}%

\biboptions{authoryear}

\begin{document}
\begin{frontmatter}

\title{Reinforcing Medical Image Classifier\\ to Improve Generalization on Small Datasets}

\author[one]{Walid Abdullah Al}
\author[one]{Il Dong Yun\corref{mycorrespondingauthor}}
\cortext[mycorrespondingauthor]{Corresponding author}\ead{yun@hufs.ac.kr}

\address[one]{Department of Computer and Electronic Systems Engineering,\\ Hankuk University of Foreign Studies, Yongin, South Korea}
%
%

%

\begin{abstract}
With the advents of deep learning, improved image classification with complex discriminative models has been made possible. However, such deep models with increased complexity require a huge set of labeled samples to generalize the training. Such classification models can easily overfit when applied for medical images because of limited training data, which is a common problem in the field of medical image analysis. This paper proposes and investigates a reinforced classifier for improving the generalization under a few available training data. Partially following the idea of reinforcement learning, the proposed classifier uses a generalization-feedback from a subset of the training data to update its parameter instead of only using the conventional cross-entropy loss about the training data. We evaluate the improvement of the proposed classifier by applying it on three different classification problems against the standard deep classifiers equipped with existing overfitting-prevention techniques. Besides an overall improvement in classification performance, the proposed classifier showed remarkable characteristics of generalized learning, which can have great potential in medical classification tasks.

\end{abstract}


\end{frontmatter}


\section{Introduction}
\label{sec:introduction}
While the usage of deep learning in medical image classification is growing rapidly, the problem of overfitting is often being raised as a major drawback \citep{lee2017deep}. To train the deep networks effectively, large scale data are required. However, collecting labeled medical image data is difficult and the research often has to be proceeded using a limited dataset. Therefore, training the model while improving model generalization for unseen data is a major challenge in deep learning-based medical image classification. Some common workarounds include data augmentation \citep{frid2018synthetic}, transfer learning \citep{sahiner2019deep, samala2018evolutionary}, etc. However, medical image data is sensitive and fundamentally different from the natural image. Therefore, such workarounds risk using artificial medical data (in case of data augmentation) which may affect the model in unpredictable ways, or using representations learned from different domains (in case of transfer learning). 
\par

For learning with the original data, two basic principles are usually applied to prevent overfitting- (i) forcing robustness, and (ii) forcing model simplicity. For the first principle, the dropout technique \citep{srivastava2014dropout} is popular among deep neural network-based models, which suggests partial update of the model by randomly assigning some neuron-outputs to zero. For the second application, reducing the number of parameters per each layer or the number of layers is one approach. On the other hand, penalizing the weights by adding a regularization cost \citep{xie2016disturblabel} in the objective function keeping the original model is the most common approach. However, reducing model complexity may not be useful when the actual decision is complex. All these techniques are based on supervised learning, where the network is optimized only to fit the training data with no explicit control on the generalization about the unseen data.

To prevent overfitting of the deep classifier under the small dataset condition, we propose a reinforced classifier that uses a scalar feedback as a generalization hint in its objective function. This reinforcement enable the classifier to aim for the performance on the unseen data, instead of hard-matching its output to the training example using cross-entropy loss. To learn from using such indirect signal, we partially applied the idea of policy gradient methods in reinforcement  (RL) \citep{arulkumaran2017deep} because standard supervised approaches only allow to optimize based on direct matching with a known answer. We tested our method for two medical image classification problems against the standard deep classifiers. We also used a public small dataset for natural image classification. All the experiments showed a significant improvement in reducing the overfitting, thus improving the classification performance on the test data.

\section{Background}
\subsection{Deep Classification Model}
\label{sec:deep}
Among the deep learning models, convolutional neural network (CNN) \citep{krizhevsky2012imagenet} is widely adopted for image classification for its implicit hierarchical feature learning mechanism. With numerous cases of its application, this has shown significant improvement in general image classification problems \citep{gao2018sdcnn, lakhani2017deep, huynh2016digital, sahiner2019deep}. Moreover, the architectures are getting heavier to enable modeling more complex decisions. Nevertheless, the existing deep classification models usually represent a common system despite having varied internal architectures. Assuming a set of target classes $C$, such system provides a class-wise probability distribution $f(X,y) = P(y|X)$ for a given image input $X$, where $y \in C$. These models are trained using pure supervised approach. For a given labeled training set $T = (\boldsymbol{X},Y)$ with image set $\boldsymbol{X}$ and the corresponding true class label set $Y$, the goal of training is to minimize a loss function, which is defined to represent the fitness of the model output to the true distribution. The widely used loss function for softmax-gated classification \citep{goodfellow2016deep} is the cross entropy function, which is as follows:
\begin{equation}
L_f(T) = \mathbb{E}_{(X,y_t) \in T} \big[ - \log f(X,y_t) \big]
\label{eq:crosss_entropy}
\end{equation}
During model training, the associated function parameter is updated to minimize this loss function for the labeled images $(X,y_t)$ in the training set $T$. Such minimization is often performed by means of stochastic gradient descent iterating over a number of epochs. At each epoch, random minibatches sampled from the training set are used for updating the network parameters. 

Deep model with high network complexity can easily fit to the training data by performing a well approximation of the true class distribution, i.e., $f(X,y) \approx P(y|X)$ about the training set $T$. However, overfitting can arise due to such complexity when the model loses generalization while fitting to the training data, causing a poor performance on the unseen test data \citep{caruana2001overfitting}. This is a common problem in deep learning. To keep track of the generalization while updating the network, a validation set is commonly used. The validation set can be seen as a subset of the original training set, whereas the remaining subset is used as the actual training set for the network-update. The performance on both the training and validation sets are examined at each epoch. The network parameters at the epoch with the best validation set performance is chosen to be the final model parameters with an optimal generalization. Despite being the best among other epochs, the performance on the validation set at the chosen epoch can still be poor.

\subsection{Reinforcement learning}
\label{sec:rl}
In the pure supervised learning, a loss function is minimized to find a good fit to a finite set of training samples enlisting the model inputs and the corresponding true outputs. On the other hand, the goal of RL is to optimize the policy of an agent to maximize the cumulative reward achievable through a sequential Markov decision process (MDP) \citep{mnih2015human}. The agent at a certain state $s$ in an environment, and decides an action $a$ based on the observed state, to update to a new state $s'$. The agent repeatedly goes through such transitions for a predefined episode. After each transition the agent receives a reward signal $r$. The transition along with the reward is considered as the experience of the agent $e=(s,a,s',r)$. The learning objective is formulated as to find an optimal behavior (or policy) $\pi(s,a)$ which maximizes the cumulative return of an episode. Here, policy $\pi(s,a)$ gives the optimal action probability for a given state $s$. 

In deep RL, the policy is parameterized using a deep neural network. The objective function to learn the optimal policy parameter using the simplest policy based RL can be expressed as:
\begin{equation}
J_\pi\big((s,a,s',r)\big) = \mathbb{E}_{(s,a,s',r)} \big[ r \log \pi(s,a) \big]
\label{eq:policy}
\end{equation}
Policy can also be seen as an action-classifier. However, unlike the finite and predefined training samples with known output labels as in the supervised learning, the experience $(s,a,s',r)$ used for training the policy are gathered by the agent. Moreover, the policy is optimized to maximize rewards instead of fitting its output directly to the correct action by minimizing the loss between the predicted action distribution and some known true distribution. Thus, RL allows to optimize a model output that may not have a direct and immediate influence on the actual problem to be solved. On the other hand, supervised learning performs a direct matching of the output against the known answer to the problem. Therefore, training is difficult and slow in case of RL.

An advantage function $A(s,a,s',r)$ is often used instead of using only the reward $r$ in the right hand side of \eqref{eq:policy} to have a reduced variance in learning \citep{schulman2017proximal}. To calculate the advantage for an experience $(s,a,s',r)$, a discounted return $r_{\gamma}=r+\gamma v(s')$ is calculated where $v(s)$ represents the value function approximating the cumulative future return achievable through state $s$ using the current policy. Finally, the advantage is calculated as: $ A(s,a,s',r) = r+v(s')-v(s) $, indicating the advantage of the explored transition compared to the discounted return provided by $v(s)$ approximated for the recent policy. At each iteration, value $v$ is also updated to approximate $r_{\gamma}$ according to the current experience.

\section{Reinforced Classifier}
To improve model generalization under the condition of few data, we equip the standard deep classifier with a generalization feedback. The proposed reinforced classifier allows to learn complex model for the medical image classification, however, without losing generalization. For such learning, we redefine the learning goal and process of the classification model, which is described in the following subsections. 

\subsection{Learning objective}
\label{sec:objective}
As discussed in Subsection~\ref{sec:deep}, the usual approach for estimating generalization is based on a validation set, which is a small subset of the training set. In this approach, the original training set is randomly split into two exclusive subsets- the active training set $T$ and the validation set $V$. During training, the model is optimized to have a minimal loss about $T$. On the other hand, the performance on the validation set $V$ is used to obtain an estimation about the generalization of the model about the unseen data. Though the supervised learning objective is to minimize the training loss, the ultimate goal is to look for the best validation performance to choose the final generalized model. However, this performance is merely observed during training instead of being controlled. We suggest a way to control the validation performance while still operating on the training data.

Model-update in the supervised learning moves towards matching the output for a given training image to the corresponding true class. We emphasize that such update has long-term future consequence in the generalization of the model beyond mimicking the training examples. We propose to embed such consequence into our objective function. Using the validation set performance as the quantification of generalization, our goal is to update the model to maximize the validation set performance. We use the cross-entropy loss as the performance measure. Taking the negative of the cross-entropy loss, we can still formulate our objective as a maximization problem.

\subsection{Learning framework}
\label{sec:framework}
With the above redefined objective, it is difficult to learn the model using supervised approach. This is because supervised learning optimizes the model by comparing its predicted output to the known true output, where the model prediction and the true target are of the same kind (in this case, class-distribution). On the other hand, the model prediction (i.e., class-distribution) is different from the target (i.e., validation performance) in the proposed classifier. It is not possible to compare these two. The target validation performance can only be described as a future consequence of the model-update. For example, if the model is updated towards a certain class for a given training image, this update would also influence the validation performance.

\begin{figure}[h]
\centering\includegraphics[scale=0.8]{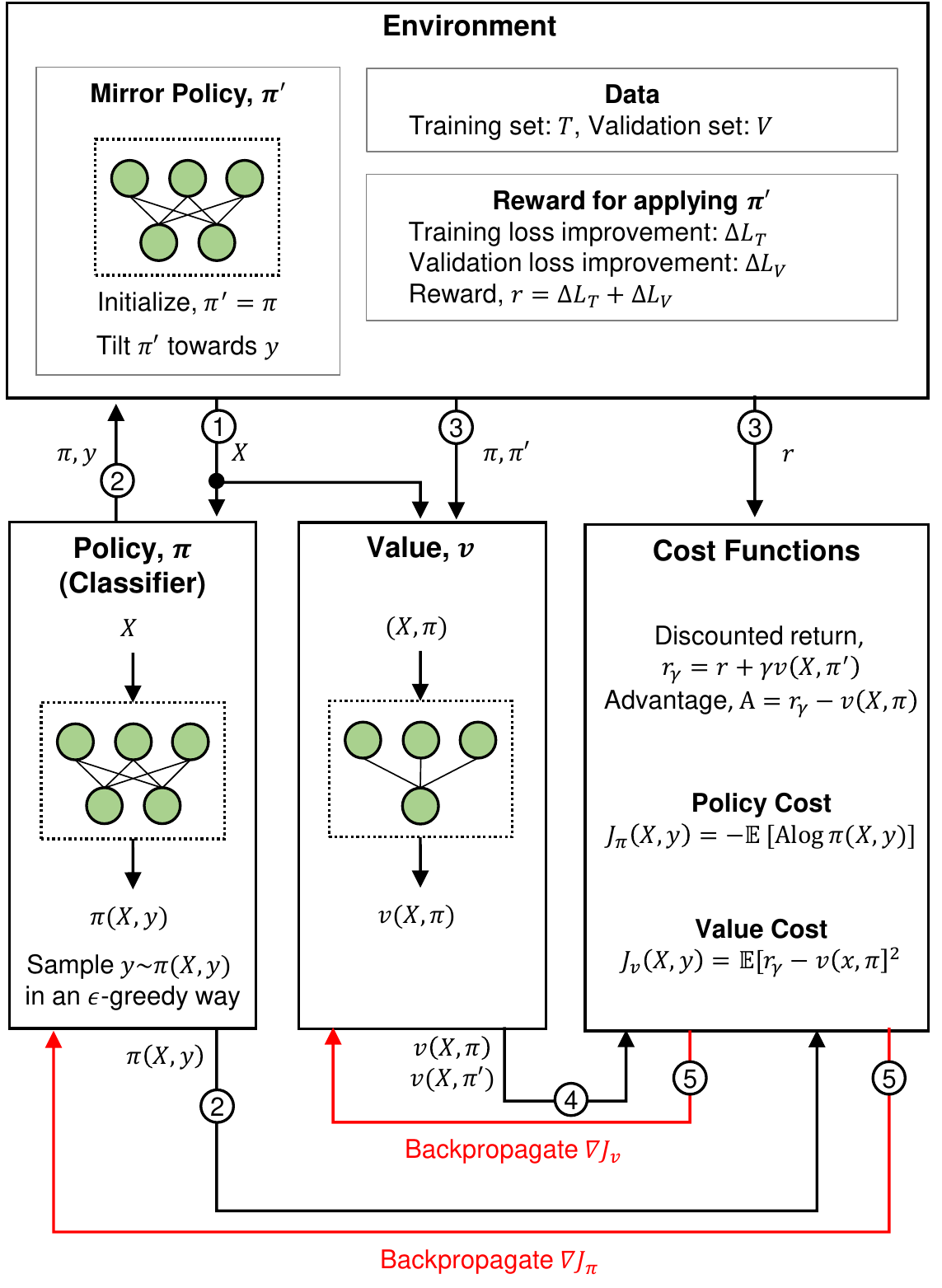}
\caption{{\bf Framework of the proposed reinforced classifier.} The agent observes a training image $X$, chooses a class $y$ based on its policy $\pi(X,y)$. It then \textit{tilts} the mirror policy so that it gives more probability towards $y$ for input $X$. The reward is provided based on the training and validation loss improvement caused by such \textit{tilting}. Using this reward, the agent updates its policy for the chosen class accordingly.}
\label{fig:method}
\end{figure}

RL can optimize a model only looking at some future feedback (i.e., reward), as discussed in Subsection~\ref{sec:rl}. Therefore, we propose to follow the RL framework to attain the proposed learning objective. Thus, we formulate the generalization problem as the behavioral task of an RL agent. For a given image, the agent tries to update its policy towards a certain class, in order to maximize the long-term validation performance. To apply the RL framework onto our problem, we must relate our problem to the key RL elements- state, action, reward, policy and value. Figure~\ref{fig:method} illustrates the overall framework of the proposed classifier.

\textbf{State} is simply the training image $X$ that the agent observes to decide an action. \textbf{Action} is sampled from the possible output classes. Therefore, we denote the action by class label $y$. \textbf{Policy} $\pi(X,y)$ can be referred to as the classification model giving class-wise distribution $P(y|X)$. Applying action $y$ for a given state $X$ suggests \textit{tilting} or updating the classifier (or, the policy) towards $y$, or in other words, editing the policy so that it outputs an increased probability of $y$ for input $X$. \textbf{Reward} $r(X,y)$ can be defined as the improvement in generalization caused by applying action $y$ for a given state $X$. Defining \textbf{value} about the proposed state is not useful because there is no transition process or sequential relation between states. Therefore, only for the value function, we use an augmented state $(X,\pi)$, which is obtained by pairing the policy with the original state. Therefore, we denote the value function by $v(X,\pi)$. Intuitively, $v(X,\pi)$ indicates the value of applying policy $\pi$ for state $X$. This augmented state can be used to represent a transition of a one-step MDP, where $(X,\pi)$ is transitioned to $(X, \pi ')$ by taking action $y$. Here, $\pi$ is the original policy and $\pi'$ is the \textit{tilted} policy obtained for action $y$. Practically, we obtain this augmented state by concatenating the feature vector after the last convolutional layer of the policy and the output class-distribution (i.e., the final layer output of the policy).

In RL, states are provided by the environment. While the agent interacts by applying action on the environment to reach a new state, the environment returns the corresponding reward. In our scheme, the actions are applied on the policy itself. Therefore, policy also becomes a part of the environment. Also, the agent models a self-editing policy where it generates optimal action by its policy and applies the action to again edit the policy. However, once the policy is edited for a state to explore the validation performance improvement, using the changed policy during exploration of another state is not fruitful, considering that the states are not sequential. Therefore, we keep a mirror policy $\pi'$ inside the environment which imitates the original policy $\pi$, and apply \textit{tilting} actions on the mirror policy keeping the original policy untouched during exploration.

\subsection{Learning Process}
\label{sec:process}
Because the states are already known and are not explored by a sequential decision process, we cannot fully follow the process of the RL. We describe the learning process as the epoch progression of standard supervised learning approach. Here, each epoch has two major phases-exploration and update. We denote the trainable parameters of the policy, mirror policy, and value by $\Theta_{\pi}$, $\Theta_{\pi'}$, and $\Theta_{v}$

\subsubsection{Exploration Phase}
In this phase, the agent gathers experience by exploring the environment. As the states (i.e., training images) are fixed, it explores different actions for each of the training images. For each state $X$, it samples an action (i.e., class) $y$ based on the policy $\pi(X,y)$ and \textit{tilts} the policy towards the class $y$. The action is chosen using an $\epsilon$-greedy approach \citep{wunder2010classes}, where a greedy action indicated by the policy is chosen with a probability of $\epsilon$ and a random action is chosen with a probability of $1-\epsilon$. To have an increased probability for the chosen class $y$, we \textit{tilt} the mirror policy $\pi'$ to increase the log likelihood of $y$ for the current state $X$. We do so by backpropagating the necessary gradients through the mirror policy network to update its parameter \citep{hecht1992theory}. The update rule can be expressed as:
\begin{equation}
\label{eq:mirror}
\Theta_{\pi'} \leftarrow \Theta_{\pi'}+\alpha_{\pi'} \nabla_{\Theta_{\pi'}} \log \pi'(X,y)
\end{equation}
where $\alpha_{\pi'}$ is the \textit{tilting} rate. Note that, the parameter of the mirror policy $\Theta_{\pi'}$ is initialized to the actual policy parameter $\Theta_{\pi}$ at the beginning. The reward is calculated in terms of the loss improvement caused by this \textit{tilting}. Though our major goal is to improve the validation loss, we also use the training loss improvement in the reward function. It is difficult to correlate the training data to the validation loss only, causing a difficult learning. Therefore, we add the training loss information in the reward. The proposed reward is as follows:
\begin{equation}
\label{eq:reward}
r(X,y) = \Delta L_T + \Delta L_V
\end{equation}
where
\begin{equation*}
\begin{split}
\Delta L_T &= L_{\pi} (T) - L_{\pi'} (T)\\
\Delta L_V &= L_{\pi} (V) - L_{\pi'} (V)
\end{split}
\end{equation*}
Finally, we record the $v(X,\pi)$ and $v(X,\pi')$ after traversing the value network for both the policies. Thus, storing such five elements of experience (i.e., training image $X$, chosen action $y$, reward $r$, value at the original policy $v(X,\pi)$ and value at the tilted policy $v(X,\pi')$), we move to the update phase. We denote this experience set as $E$.

\subsubsection{Update Phase}
Here, we update the policy and value networks based on the experiences gathered in the exploration phase to maximize the cumulative reward. Similar to the stochastic gradient descent approach in supervised learning, we randomly sample minibatches from the stored experiences and perform batch-wise updating of the networks. For each minibatch of experiences, we calculate the advantage of the \textit{tilted} policy compared to the original one. Assuming an experience $\big( X,y,r,v(X,\pi),v(X,\pi')\big)$, we first calculate the discounted cumulative return for the \textit{tilted} policy as: $r_{\gamma} = r + \gamma v(X,\pi')$, then compute the advantage compared with the one for the original policy as indicated by the value function, which is as follows:
\begin{equation}
\label{eq:advantage}
A = r_\gamma - v(X,\pi)
\end{equation}
Using this advantage, we obtain the cost function following \eqref{eq:policy}. We use the negative of the objective function in \eqref{eq:policy} to allow for the well-known minimization-based stochastic gradient descent optimization. The cost for value function is obtained by using the squared difference of the current value output $v(X,\pi)$ and discounted cumulative return $r_\gamma$ for the \textit{tilted} policy. Thus, we express the update rules for both the policy and value networks based on the cost functions as follows:
\begin{equation}
\label{eq:update}
\begin{split}
\Theta_\pi &\leftarrow \Theta_\pi + \alpha_{\pi} \nabla_{\Theta_{\pi}} \mathbb{E}_{(X,y)\in E} \big[A \log \pi(X,y) \big]\\
\Theta_v &\leftarrow \Theta_v - \alpha_{v} \nabla_{\Theta_{v}} \mathbb{E}_{(X,y) \in E} \big[ r_\gamma - v(X,\pi) \big]^2
\end{split}
\end{equation}
where $(X,y)$ is sampled from the current minibatch. 

By performing such batch-update over a number of epochs, the classification model (i.e., policy) can be trained while maintaining the generalization. However, the training progress becomes slow because the formulated reward provides a weak supervision to guide the training. To speed up the training, we perform an extra step of batch-update on the policy for each minibatch, where the model is updated based on the original supervised cross-entropy loss about the training set. If we combine this update with the reinforced policy update in \eqref{eq:update}, we can present the overall policy update as follows:
\begin{equation}
\label{eq:policyupdates}
\begin{split}
\Theta_\pi \leftarrow \Theta_\pi + &\alpha_{\pi} \nabla_{\Theta_{\pi}} \mathbb{E}_{(X,y)\in E} \big[A \log \pi(X,y) \big]\\
+ &c\alpha_{\pi} \nabla_{\Theta_{\pi}} \mathbb{E}_{X\in E} \big[A \log \pi(X,y_t) \big]
\end{split}
\end{equation} 
Note that, $y_t$ in the supervised update part is the true class label of experienced training image $X$ obtained from $T$. The learning rate for the supervised update part is dampened by $c
<1$. We summarize the whole training procedure in Algorithm~\ref{alg:training}.

\begin{algorithm}[h]
\caption{Learning process of the reinforced classifier}
\label{alg:training}
\begin{algorithmic}
\STATE {\bf Input}
\STATE Training set $T$, Validation set $V$
\STATE Learn-rate for policy, value, and mirror policy: $\alpha_\pi, \alpha_v, \alpha_{\pi'}$
\STATE Maximum number of epochs $K$
\STATE {\bf Output}
\STATE Policy (i.e., classifier) $\pi$ with parameter $\Theta_{\pi}$
\STATE Value $v$ with parameter $\Theta_{v}$
\STATE {\bf Initialize} $\Theta_{\pi}, \Theta_{v}$

\FOR {$k=1$ to $K$}
\STATE Experience, $E=\emptyset$
\FOR {each image $X \in T$}
	\STATE Sample class $y\sim\pi(X,y)$ using $\epsilon$-greedy way
	\STATE Update mirror policy, $\Theta_{\pi'}\leftarrow\Theta_{\pi}$ 
	\STATE Tilt mirror policy towards $y$ using \eqref{eq:mirror}
	\STATE Compute reward $r$ for applying $\pi'$ using \eqref{eq:reward}
	\STATE Obtain value for the original policy, $v^0=v(X,\pi)$
	\STATE Obtain value for the tilted policy, $v' = v(X,\pi')$
	\STATE Append $(X,y,r, v^0, v')$ to $E$
\ENDFOR
\STATE Sample $N$ minibatches $\mathbb{B}=\{B_1,B_2,...,B_N\}$ from $E$
\FOR {each minibatch  $B\in \mathbb{B}$}
	\STATE Compute advantage $A=r+v'-v^0$ for each experience in $B$
	\STATE Update $\Theta_v$ using \eqref{eq:update}	
	\STATE Update $\Theta_\pi$ using \eqref{eq:policyupdates}

\ENDFOR
\ENDFOR
\end{algorithmic}
\end{algorithm}

\section{Results and Discussion}
\subsection{Dataset and Experiment}
For investigating the generalization performance of the proposed medical image classifier under few training data, we used two small medical image datasets. Additionally, we performed a comparative analysis using a public natural image dataset having overfitting characteristics, to strengthen the comparison in general. The first medical image dataset consists of $100$ CT images of the vermiform appendix, aimed for acute appendicitis diagnosis (i.e., classification between acute appendicitis and non-appendicitis). This dataset was also used in \citep{kim2012low}. The second dataset has $60$ MR images of breast cancers aimed for classifying the cancer subtypes- luminal-A (LA), luminal-B (LB), and human epidermal growth factor receptor-2 (HER-2). The region-of-interst (ROI) information for these two medical image datasets were obtained from the corresponding clinical sites. The public natural image dataset \citep{lazebnik2005maximum} used here consists of $300$ images of three birds ($100$ images per bird). The bird images are of different sizes. Therefore, we resized them to have a fixed input size, while maintaining the original aspect ratio. Table~\ref{tab:dataset} provides the description of the datasets including the input image/patch size. All the three datasets have limited data, and therefore, are good examples of overfitting when trained using deep CNN. 
\par

\begin{table*}[]
\caption{Description of the classification datasets.}

\centering
\renewcommand{\arraystretch}{1.5}
\setlength{\tabcolsep}{5pt}
\begin{tabular}{l l M}
\hline
{\bf Dataset} & \makecell[tl]{{\bf Image-size}\\\relax [pixels/voxels]}
& \makecell[tl]{{\bf Class-distribution}\\\relax [class-name (count)]}\\
\hline
\makecell[tl]{Birds \citep{lazebnik2005maximum}\\\relax (public)} & $64\times 64\times 3$ 
& \makecell[tl]{Egret (100)\\Mandarin duck (100)\\Snowy owl (100)}\\ 
Appendicitis \citep{kim2012low} & $100\times 100\times 30$
& \makecell[tl]{Acute appendicitis (32)\\Non-appendicitis (68)} \\ 
Breast Cancer & $80\times 80\times 7$
& \makecell[tl]{Luminal-A (20)\\Luminal-B (20)\\HER-2 (20)}\\
\hline
\end{tabular}

\label{tab:dataset}
\end{table*}

Besides implementing the proposed reinforced model on these three datasets, we also implemented the well-known overfitting-prevention approaches such as L2-regularization \citep{xie2016disturblabel}, and dropout \citep{srivastava2014dropout}, for comparison. The underlying CNN architectures for all these approaches (including ours) are identical. The optimal hyper-parameters for different approaches were determined by trial-and-error process. The network description and the hyperparameters are mentioned in \ref{apx:network} and \ref{apx:param}. The datasets were randomly split into training, validation, and testing sets with a respective ratio of 3:1:1. An identical dataset-split was used for all the methods under comparison.

\subsection{Learning Characteristics}
\begin{figure*}[!t]
\centering
\renewcommand{\arraystretch}{1.5}
\setlength{\tabcolsep}{1pt}
\begin{tabular}{l l l}
\begin{tikzpicture}
\pgfplotsset{%
    width=0.3\textwidth,
    height=0.3\textwidth,
}
\begin{axis}[
     x label style={at={(axis description cs:0.5,0.0)},anchor=north},
    y label style={at={(axis description cs:0.2,.5)},anchor=south},     
    xlabel={Epochs},
    ylabel={Classification accuracy (\%)}, 
    ymin=30, ymax=100,
    xtick={0,600,1200},
    ytick={0,20,40,60,80,100},
    legend pos=south east,
]

\addplot [mark=none,color=ao] table [col sep=comma, mark=none] {rc_tr.csv};
\addplot [mark=none,color=blue] table [col sep=comma, mark=none] {rc_val.csv};
\addplot [mark=none,color=red] table [col sep=comma, mark=none] {rc_tst.csv};
\end{axis}
\node[text width=2.5cm, align=center] at (1.65,-1.6) {(a) Reinforced};
\end{tikzpicture} &
 \begin{tikzpicture}
\pgfplotsset{%
    width=0.3\textwidth,
    height=0.3\textwidth,
}
\begin{axis}[
     x label style={at={(axis description cs:0.5,0.0)},anchor=north},
    y label style={at={(axis description cs:0.05,.5)},anchor=south},     
    xlabel={Epochs},
    ylabel={ },    
    xmin=0, xmax=420,
    ymin=30, ymax=100,
    xtick={0,200,400},
    yticklabels={,,},
    legend pos=south east,
]

\addplot [mark=none,color=ao] table [col sep=comma, mark=none] {drop_tr.csv};
\addplot [mark=none,color=blue] table [col sep=comma, mark=none] {drop_val.csv};
\addplot [mark=none,color=red] table [col sep=comma, mark=none] {drop_tst.csv};

\end{axis}
\node[text width=2.8cm, align=center] at (1.65,-1.6) {(b) Dropout};
\end{tikzpicture}
%
%
&
\begin{tikzpicture}
\pgfplotsset{%
    width=0.3\textwidth,
    height=0.3\textwidth,
}
\begin{axis}[
     x label style={at={(axis description cs:0.5,0.0)},anchor=north},
    y label style={at={(axis description cs:0.05,.5)},anchor=south},     
    xlabel={Epochs},
    ylabel={ },    
    ymin=30, ymax=100,
    xtick={0,600,1200},
    yticklabels={,,},
    legend pos=south east,
]

\addplot [mark=none,color=ao] table [col sep=comma, mark=none] {dropl2_tr.csv};
    \addlegendentry{Train}
\addplot [mark=none,color=blue] table [col sep=comma, mark=none] {dropl2_val.csv};
    \addlegendentry{Validation}
\addplot [mark=none,color=red] table [col sep=comma, mark=none] {dropl2_tst.csv};
    \addlegendentry{Test}

\end{axis}
\node[text width=2.8cm, align=center] at (1.65,-1.6) {(c) Dropout+L2};
\end{tikzpicture}
\end{tabular}

\caption{{\bf Learning curves of the proposed approach against the other overfitting-prevention approaches.} The reinforced approach could notably reduce the generalization gap.}
\label{fig:learning}
\end{figure*}
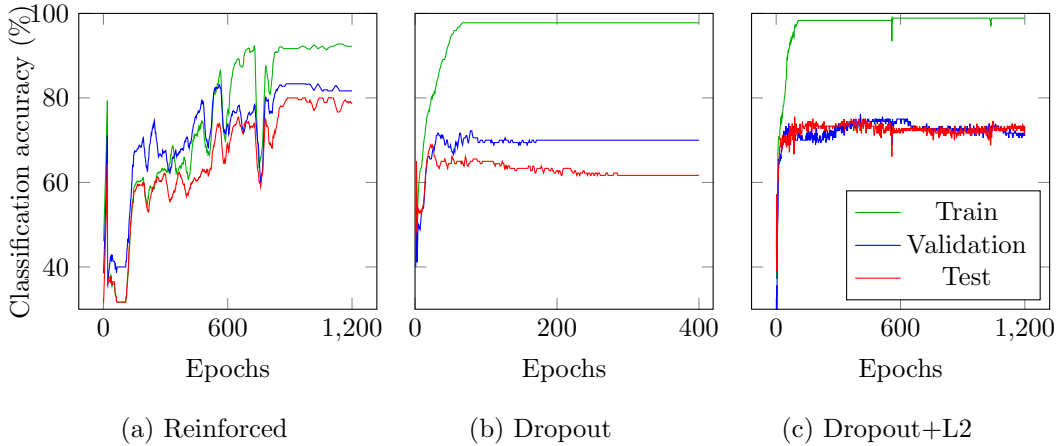

To present the comparative learning characteristics, we plot the accuracy of the training, validation and test sets over the epochs in Figure~\ref{fig:learning}. In this figure, we also plot such learning progress for the other overfitting-prevention approaches mentioned above. We allowed running the training until the accuracy converges or the validation accuracy falls consistently without showing any sign of improvement. The test accuracy was not used in any decision, therefore, treated as entirely unseen. The presented plots are for the birds dataset. The other two datasets also showed a similar pattern of learning curves. Therefore, we chose not to include them to avoid enlarging the figure redundantly.

The overfitting situation with such small dataset is clear in the learning curves in Figure~\ref{fig:learning}. Even the dropout technique together with regularization could cause only a slight improvement in generalization. However, the generalization gap (i.e., the gap between training and test/validation curves) is significantly small in the proposed approach, compared with the other approaches. The test/validation curves could better follow along with the training curve in this case, causing a reduced generalization error. The existing supervised approaches, despite the constraints (e.g., partial network update in case of dropout, and weight penalizing in case of regularization) solely work on improving the training accuracy over the epochs. The accuracy for the test and validation sets are simply byproducts of the training improvement, to which the training has no explicit contribution. Therefore, after a certain period, these performances no longer improve with the training. On the other hand, there are two major network updates at each epoch in the proposed method- the supervised update aiming at maximizing the training set performance, and the reinforced update attempting to maximize the validation set performance. Because of this competition, the training accuracy cannot easily improve without considering the validation accuracy, while the training accuracy in the other approaches follows a smooth increase comparatively. Consequently, all the curves are forced to follow a similar path reducing the gap in between.

The pure supervised update is based on the likelihood of the true class only. In our approach, the reinforced update running in parallel to the supervised update, conducts constant exploration of different classes to obtain the maximized validation performance. Consequently, the learning is comparatively slow and goes through high alteration in performance, despite giving a harmonized progress in the training and validation performance. Such high amount of alteration is usual in RL.

\subsection{Generalization}
Though we have used the validation set performance to have an idea about the generalization, this estimate does not represent the actual generalization on the unseen test data. Considering the test set entirely unseen, we obtain the final network from the epoch where the validation performance is the maximum. We present the training, validation, and test set performance of such epoch in Table~\ref{tab:optimal_epoch}. The proposed approach showed significant improvement in generalization error on the test set. For all the datasets, the improvement is apparent compared with the other approaches. While the dropout combined with regularization gave better generalization than the dropout-only approach, the proposed approach showed about 11\% further improvement on average in all the cases, despite using the same training data. The major difference of the proposed approach is that it conducts an additional reinforced update based on the validation feedback. Such update allows the classifier to learn to generalize as opposed to learning to fit the training data only. While operating only on the seen training data, it tries to optimize a scalar feedback about some unseen data while not actually looking at those data, thus encoding the generalization behavior in its learning process. Thereby, despite having trained for validation performance, the trained classifier shows improved generalization on the actual test data.

\begin{table*}[]
\caption{Classification error (\%) of the training, validation (Val.) and testing subsets at the optimal epoch.}

\centering
\renewcommand{\arraystretch}{1.2}
\setlength{\tabcolsep}{2.5pt}
\begin{tabular}{l  c c c  m c c  m c c}
\hline
\multicolumn{1}{l}{\multirow{2}{*}{\bf Method}} & \multicolumn{3}{c}{\bf Birds} &
\multicolumn{3}{m}{\bf Appendicitis} &
\multicolumn{3}{m}{\bf Breast Cancer} \\
& \makecell{Train} & \makecell{Val.} & \makecell{Test}
& \makecell{Train} & \makecell{Val.} & \makecell{Test}
& \makecell{Train} & \makecell{Val.} & \makecell{Test}\\ 
\hline
Reinforced & 7.22 &  16.67 & {\bf 18.33} & 11.67 &  15.00 & {\bf 15.00} & 5.56 &  8.33 & {\bf 16.67}\\
Dropout & 2.22 & 26.67 & {\bf 35.00} & 11.67 & 40.00 & {\bf 35.00} & 8.33 & 33.33 & {\bf 41.67} \\
Dropout+L2 & 1.67 & 23.33 & {\bf 28.33} & 3.33 & 35.00 & {\bf 30.00} & 2.78 & 25.00 & {\bf 25.00} \\
\hline 
\end{tabular}
\label{tab:optimal_epoch}
\end{table*}

Finally, we obtained ten different random splits of the datasets (into train, validation and test subsets). We repeated the experiments for each split to have a more general estimation of the comparative performance. We present the resultant classification error distribution for all the methods in Table~\ref{tab:monte_carlo}. Here, we present the error on the test data as the final evaluation measure. We put the error distribution of the dropout combined with L2-regularizaiton approach for comparison, because it showed a better generalization than the dropout-only approach. In general, the proposed method showed an improved performance over the existing approaches in all the datasets. On average, it has reduced the classification error by 12.50\% for the birds dataset, 8.50\% for the appendicitis dataset, and 9.17\% for the breast cancer dataset. The improvement is statistically significant showing a $p$-value less than $0.001$ for the birds and appendicitis datasets, whereas the $p$-value for the breast cancer dataset is less than $0.02$.

\begin{table*}[]
\caption{Distribution of classification error (\%) for the test data in different dataset-splits}

\centering
\renewcommand{\arraystretch}{1.5}
\begin{tabular}{l  M c  M c}
\hline
\multicolumn{1}{l}{\multirow{2}{*}{\bf Dataset}} & \multicolumn{2}{c}{\bf Reinforced} &
\multicolumn{2}{c}{\bf Dropout+L2} \\
& Mean$\pm$SD & Median & Mean$\pm$SD & Median\\
\hline
Birds & $17.33\pm 1.79$ & $16.67$ & $29.83\pm 3.46$ & $28.33$\\
Appendicitis & $22.00\pm 3.49$ & $20.00$ & $30.50\pm 4.38$ & $30.00$\\
Breast Cancer & $18.33\pm 7.66$ & $16.67$ & $27.50\pm 7.91$ & $25.00$\\
\hline
\end{tabular}

\label{tab:monte_carlo}
\end{table*}

\subsection{Saliency Map}
\begin{figure*}[]
\centering
\includegraphics[width=0.7\textwidth]{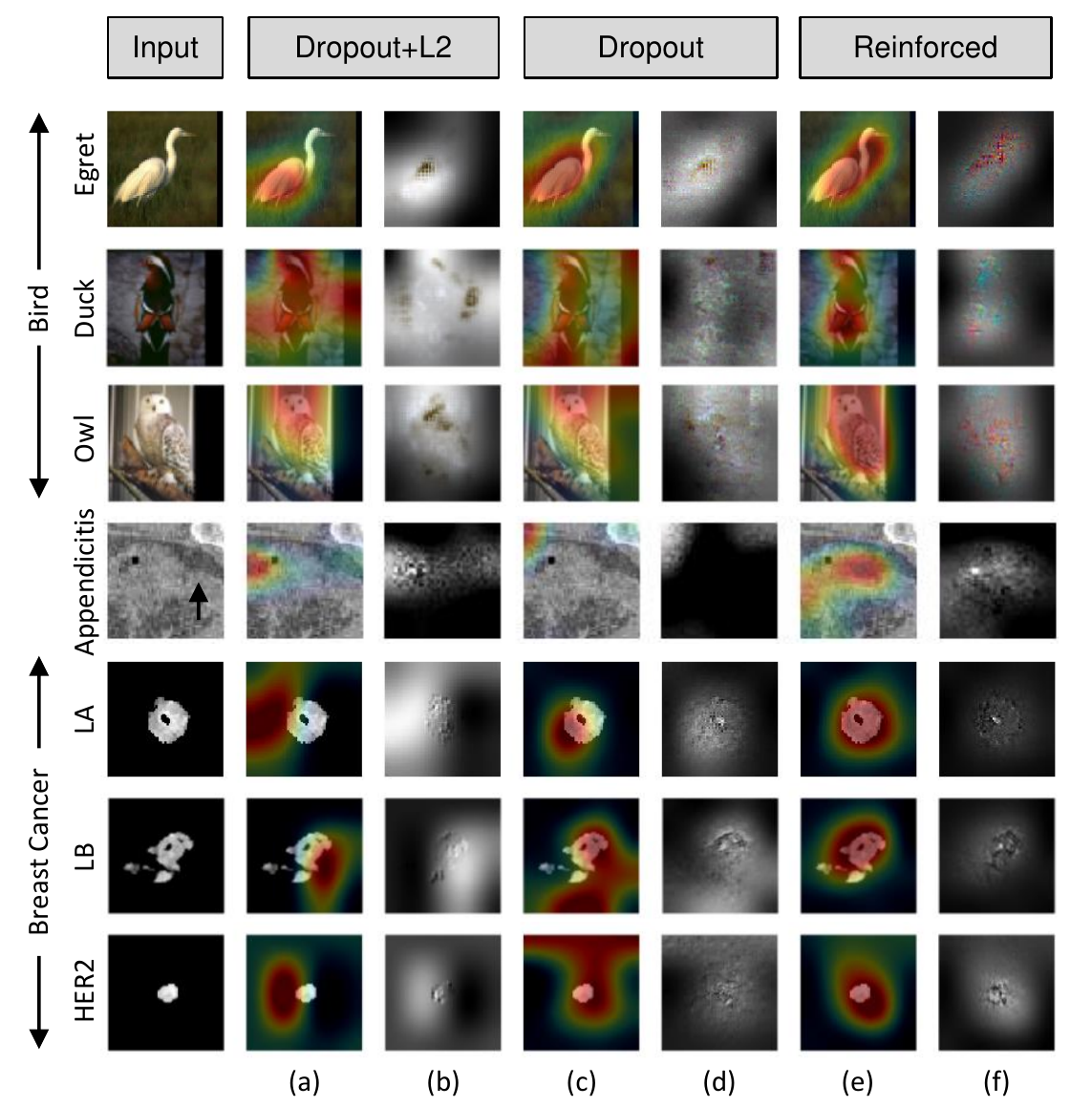}
\caption{{\bf Example saliency maps for different classes.} (a, c and e) Grad-CAM heatmaps for the corresponding classes mentioned at the left of the input images. (b, d and f) pixel-level explanations obtained by the guided Grad-CAM. The arrow indicates the appendix. }
\label{cam}
\end{figure*}

Saliency map allows to understand class-specific important features or regions as learned by a model. Therefore, analyzing such map is a powerful way to evaluate interpretability and confidence of a model. We use the gradient-weighted class activation mapping (Grad-CAM) \citep{selvaraju2017grad} to present the responsible regions in an image for deciding a class. Also, we use the guided Grad-CAM \citep{selvaraju2017grad} to obtain pixel-level visual explanation specific to a class. Figure~\ref{cam} presents the Grad-CAM and guided Grad-CAM results for inputs of different classes. 
\par

Overall, the proposed method showed stronger attention (as observed in the Grad-CAM results) to the target objects or organs compared with the other methods. From the pixel-level saliency maps of the bird-images, it is apparent that the proposed method could learn visually discriminative representation from such a small training set. The saliency maps of the proposed model hold significantly better explainability than those of the dropout and regularization-based approaches. 
\par

We could also observe improved attention for the medical image datasets. Acute appendicitis is described as the inflamed condition of the appendix. Therefore, a good appendicitis classifier should draw its attention towards the appendix. From such a few training images, the dropout and regularization approaches could not be able to focus on the appendix for making the decision. On the other hand, our reinforced approach could enable learning a decision based on the appendix region, resulting in a heatmap highlighting the appendix. Heatmaps for classifying the breast cancer subtypes also indicate better concentration on the lesions compared with the supervised approaches. Therefore, the proposed approach showed improved learning potential on the small datasets.
\par

\section{Conclusion}
\label{sec:conclusion}
To tackle the overfitting problem under limited dataset, we propose to reinforce the standard deep classifier with a generalization-feedback. Besides the supervised update to fit the training data, we conduct a reinforced update to maximize a scalar feedback representing the performance on the validation set which is a small subset of the training set. Optimizing the classifier to maximize such feedback looking only at the training data helps learn a generalization behavior instead of improving the training set performance only. Experiment with three datasets generally showed improved generalization compared with the standard overfitting-preventing methods. Besides showing generalization error reduction at the optimal epoch, the entire learning process showed a generalized progress in all the training, validation, and test set performances. The resultant saliency map with better explainability also showed promising learning potential on small training sets.

\section*{Acknowledgement}
This research was supported by Basic Science Research Program through the National Research Foundation of Korea (NRF), funded by the Ministry of Education, Science, Technology (No. 2017R1A2B4004503).

\appendix

\section{CNN Architecture}
\label{apx:network}
For a fair comparison, a common CNN architecture is used as the classifier in all the approaches under comparison. In the proposed framework, the policy network is as same as the classifier model. The value network is a stack of two fully connected layers built on top of the final feature vector output of the CNN (before the fully connected stack of the policy). The architecture of the classifier model consists of four convolutional layers with following rectified linear unit (ReLU) activations and max-pooling layers. The fourth max-pooling layer is followed by two fully connected layers, where the final fully connected layer is gated through a softmax function to generate class-wise probability score. Each convolutional layer has $3\times 3$ kernels with no stride, while the pooling layers have $2\times 2$ kernel with stride $2$. Note that, the network described here is used for the birds classification problem. Additional convolution-ReLU-pool stacks are inserted accordingly when the input size is bigger (in case of the appendix and breast cancer patches). Also, 3D kernels are used for the appendix patches. Breast cancer patches have a small-length third dimension. Therefore, we treated them as 2D considering the third dimension as channels of the usual color image, keeping the model simple.

\section{Hyperparameter Selection}
\label{apx:param}
All the hyperparameters are obtained through trial-and-error process. The optimal learning rate for the supervised update for all the classifiers (including the supervised update part in the reinforced classifier) is $1\mathrm{e}{-4}$. The learning rate for the reinforced policy and value update in the proposed approach is $1\mathrm{e}{-3}$. Therefore, the dampening factor of the supervised learning rate, $c=0.1$ in \eqref{eq:policyupdates}. In our implementation, the learning rates for updating policy and value, and tilting the mirror policy are same, i.e., $\alpha_\pi=\alpha_{\pi'}=\alpha_{v}$. There were no significant improvements in the performance when different rates were used. Following the common trends in existing RL-based studies \citep{mnih2015human, alansary2019evaluating}, we set the discount factor $\gamma$ to $0.9$. The value of $\epsilon$ for $\epsilon$-greedy exploration is gradually increased from $0.3$ to $0.7$ over the epochs. For the dropout approach, a keep probability of $0.5$ is used. As for the L2-regularization, we select a regularization scale $\lambda=0.1$ for birds dataset, and $\lambda=0.2$ for both the appendicitis and breast cancer datasets, after trying different values from  $1.0$ to $1\mathrm{e}{-3}$. 

\section*{References}

\bibliography{mybib}

\end{document}